\documentclass[10pt,twocolumn,letterpaper]{article}

\usepackage{iccv}
\usepackage{times}
\usepackage{epsfig}
\usepackage{graphicx}
\usepackage{amsmath}
\usepackage{amssymb}
\usepackage{tikz}
\usepackage{xcolor,pifont}
\usepackage{color, colortbl}
\usepackage{booktabs}
\usepackage{multirow}
\usepackage[linesnumbered,ruled,vlined]{algorithm2e}

\SetKwInput{KwInput}{Input}                % Set the Input
\SetKwInput{KwOutput}{Output}              % set the Output

\definecolor{COLOR_MEAN}{HTML}{f0f0f0}

\usepackage[breaklinks=true,bookmarks=false]{hyperref}

\iccvfinalcopy % *** Uncomment this line for the final submission

\ificcvfinal\fi

\begin{document}

%%%%%%%%% TITLE
\title{Towards Domain-Specific Features Disentanglement for Domain Generalization}

\author{Hao Chen\thanks{ \quad Corresponding author.}, Qi Zhang, Zenan Huang, Haobo Wang, Junbo Zhao\\
Zhejiang University\\
{\tt\small \{h.c.chen, cheung\_se, lccurious, wanghaobo, j.zhao\}@zju.edu.cn
}}

\maketitle
\ificcvfinal\thispagestyle{empty}\fi

%%%%%%%%% ABSTRACT
\begin{abstract}
Distributional shift between domains poses great challenges to modern machine learning algorithms. 
The domain generalization (DG) signifies a popular line targeting this issue, where these methods intend to uncover universal patterns across disparate distributions. 
Noted, the crucial challenge behind DG is the existence of irrelevant domain features, and most prior works overlook this information. 
Motivated by this, we propose a novel contrastive-based disentanglement method CDDG, to effectively utilize the disentangled features to exploit the over-looked domain-specific features, and thus facilitating the extraction of the desired cross-domain category features for DG tasks. 
Specifically, CDDG learns to decouple inherent mutually exclusive features by leveraging them in the latent space, thus making the learning discriminative.
Extensive experiments conducted on various benchmark datasets demonstrate the superiority of our method compared to other state-of-the-art approaches. 
Furthermore, visualization evaluations confirm the potential of our method in achieving effective feature disentanglement.
\end{abstract}

%%%%%%%%% BODY TEXT
\section{Introduction}

\begin{figure}[ht]
\begin{center}
   \includegraphics[width=1.0\linewidth]{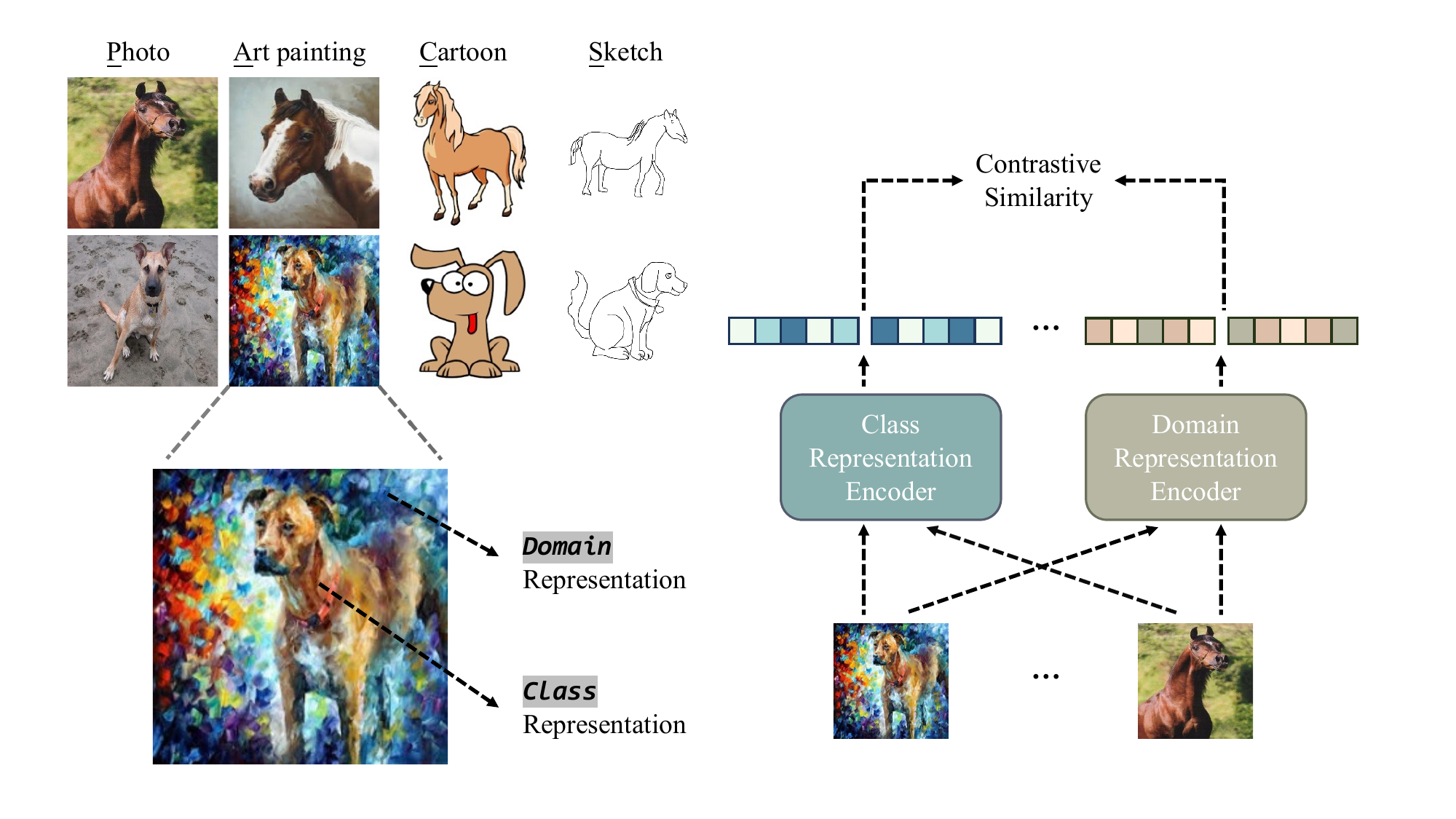}
\end{center}
   \caption{Selected images from PACS dataset, which comprises four distinct domains: Photo, Sketch, Cartoon, and Art Painting, each containing seven categories. Domain generalization aims to train models with multiple source domains and generalize to an unseen target domain, e.g., photo, art painting and cartoon as the source domains, sketch as the target domain.}
\label{fig:intro}
\end{figure}

Modern machine learning methods are primarily developed by an independent and identically distributed (I.I.D) setup in a conventional supervised learning paradigm. 
However, in real-world scenarios, the data often exhibit distributional shifts ubiquitously, posing an explicit or implicit gap between the training and inference stages~\cite{li2017deeper, gulrajani2021search}.
Domain generalization (DG) represents a line of research towards addressing this issue, with an objective rooting in uncovering the common feature among the data drawn from different domains~\cite{zhou2022domain, wang2022generalizing}.
As shown in Fig.~\ref{fig:intro}, the standardized benchmark datasets for image classification devised for DG, such as PACS~\cite{li2017deeper}, collect images of the same category from a variety of domains. 
To achieve better classification performance, with the underlying idea that data from diverse domains share a universal representation, patterns across domains that can be used for the downstream tasks, while invariant to specific domain changes, are of interest in DG.

Most prior works in DG have focused on reducing the discrepancy among embeddings from different domains, with the objective of acquiring knowledge that is invariant to domain variations, and often overlooked the utilization of domain-specific information~\cite{zhou2022domain}.
While another branch in DG focuses on decoupling features and reconstructing the original image, they do not fully utilize the decoupled features~\cite{wang2021variational}. Besides, recent work has indirectly validated that relying solely on domain representation can aid in class discrimination~\cite{cho2023promptstyler}. 
Therefore, we argue that domain-specific information can be effectively leveraged in reverse to constrain the learning of domain-invariant features. 
Furthermore, as a promising method for facilitating the learning of discriminative features, contrastive learning is compatible with the integration of feature disentanglement in the DG-specific context where mutually exclusive features inherently exist~\cite{tian2020makes, yeh2022decoupled}.
Motivated by this, a fusion of contrastive-based feature disentanglement can construct a unifying framework of representation learning for DG.

In this work, we propose a unifying feature disentanglement method called \textbf{C}ontrastive \textbf{D}isentanglement for \textbf{D}omain \textbf{G}eneralization (\textbf{CDDG}), which integrates the learning of all types of features into a unified disentanglement framework. 
Within this framework, for each type of feature, we repel all irrelevant features, including non-identical features within the same sample (as shown in Fig.~\ref{fig:fig2}(d)). 
The core idea of CDDG is to initially disentangle the features of domain generalization samples into domain features and category features, and then leverage this feature information in the latent space to assist the disentanglement process in reverse, making the inclusion of domain-specific contrastive learning serves as a constraint to facilitate domain-invariant learning.
This approach brings additional uniformity of samples/embeddings in the feature space, enabling the model to achieve better representation capability and thus improve downstream generalization performance. 
Extensive experiments conducted on various benchmark datasets demonstrate the superiority of our proposed CDDG method compared to other state-of-the-art approaches. 
Furthermore, visualization evaluations confirm the potential of our method in achieving effective feature disentanglement.

%-------------------------------------------------------------------------
\section{Related Work}

\begin{figure}[!t]
\begin{center}
   \includegraphics[width=1.0\linewidth]{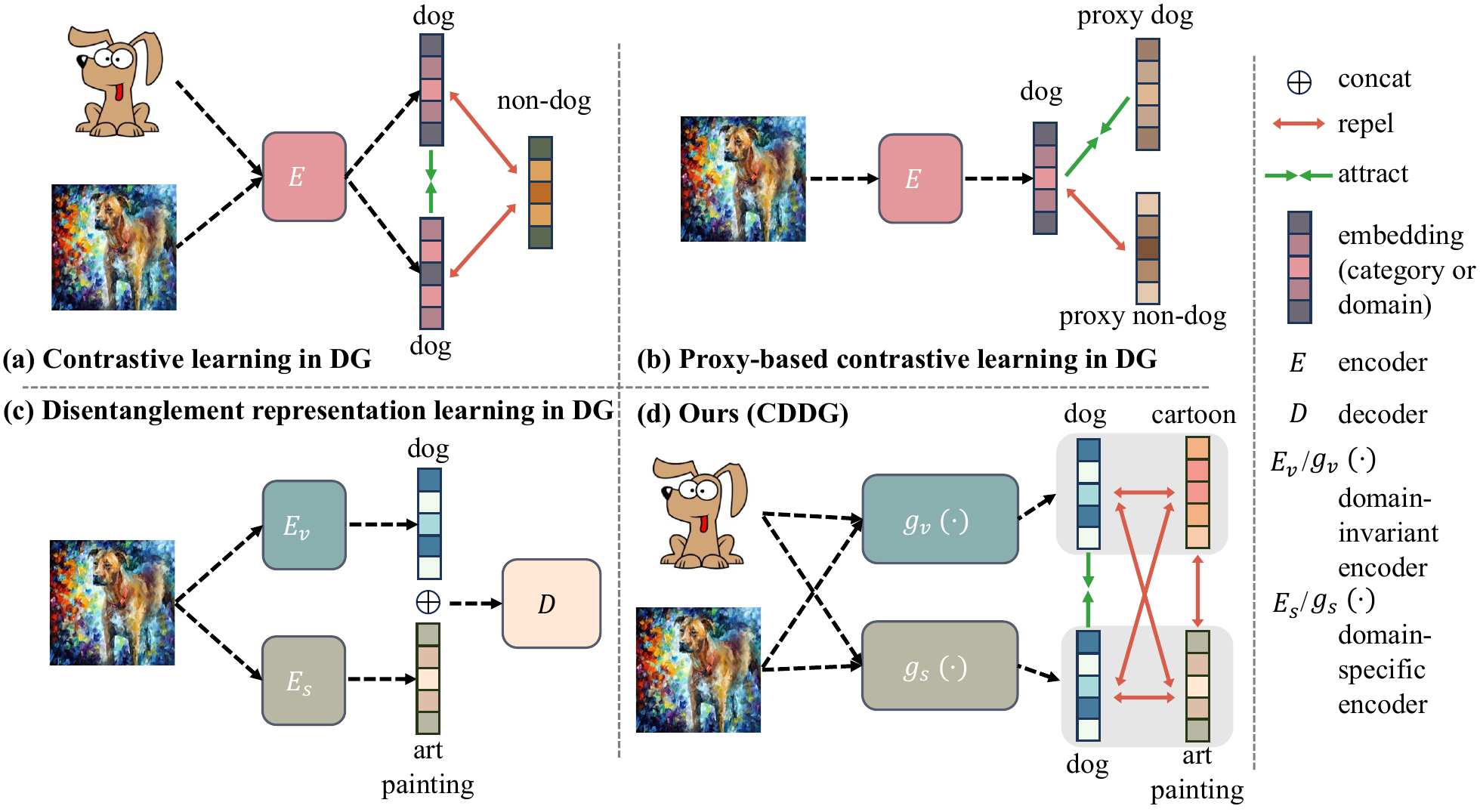}
\end{center}
   \caption{Illustration of our method compared with contrastive-based methods and disentanglement-based methods in DG. (\textbf{a}) Contrastive learning leverages features with the same category or domain. (\textbf{b}) Proxy-based CL leverages embeddings with proxies rather than embeddings of other samples. (\textbf{c}) Most disentanglement representation learning methods in DG decompose samples into domain-invariant and domain-specific features. (\textbf{d}) CDDG first decouples one sample, then leverages them in the latent space to enhance the decoupling of these features.} 
\label{fig:fig2}
\end{figure}

\begin{figure*}[!t]
\begin{center}
   \includegraphics[width=1.0\linewidth]{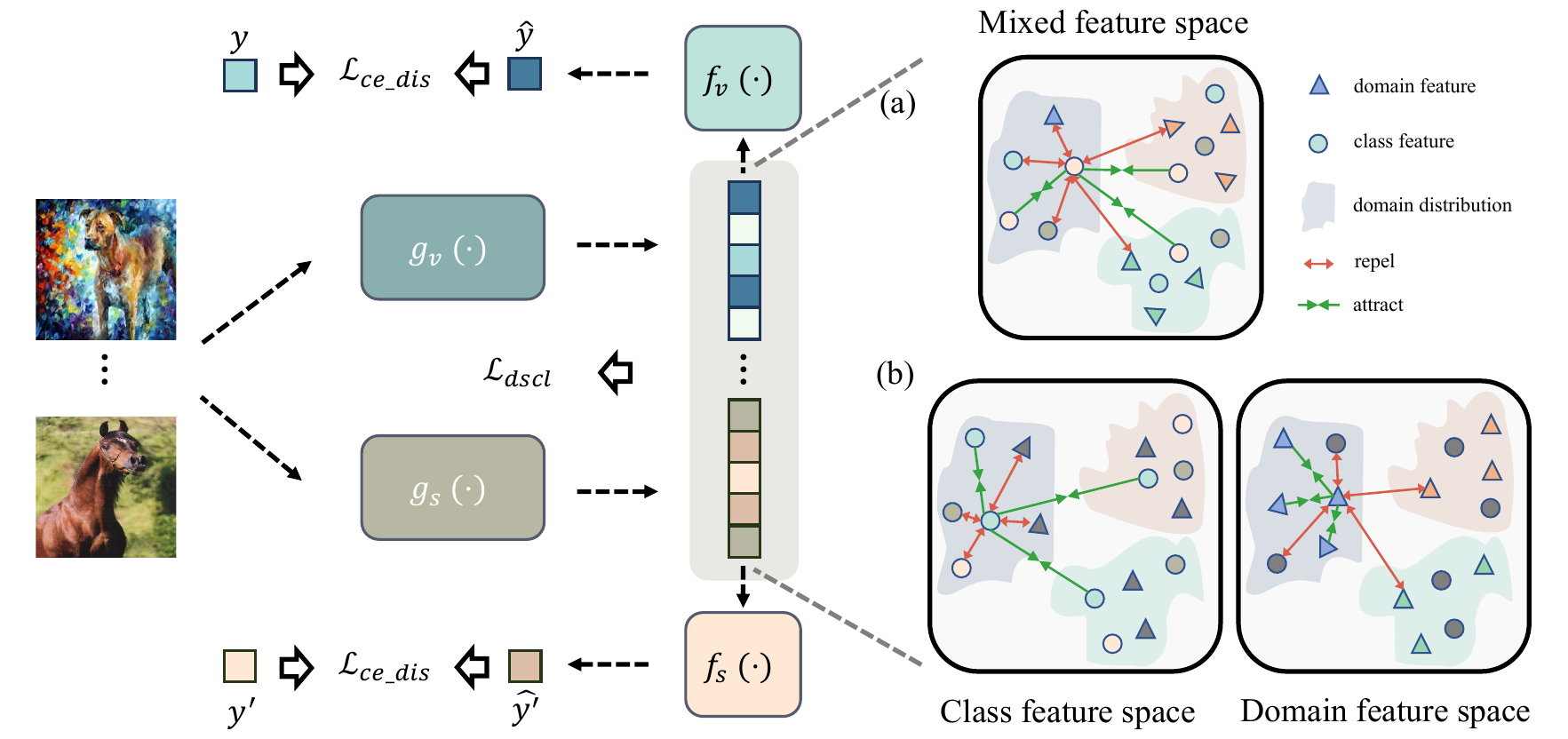}
\end{center}
   \caption{Framework of our proposed method. The input images are first fed into class feature extractor $g_v(\cdot)$ and domain feature extractor $g_s(\cdot)$. With extracted embeddings, $\mathcal{L}_{dscl}$ is then calculated. After that, class feature classifier $f_v(\cdot)$ and domain-specific feature classifier $f_s(\cdot)$ are employed to predict class label and domain label, and $\mathcal{L}_{ce\_dis}$ are then been calculated. There are two ways of mapping two features from one sample. The first is mapping both class feature and domain feature into one mixed feature space; the other is to individually map one single type of feature into one feature space, and take the other type of feature as extra negative samples. For better neatness, we have omitted a few arrows between \textit{anchor} sample with samples from different distributions. This figure is best viewed in color.}
\label{fig:method}
\end{figure*}

\paragraph{Domain Generalization.} 
Most DG methods can be categorized into three groups: 
1) Data manipulation, which mainly diversifies training data to assist in improving the model's generalization ability. This includes random flip, rotation, crop~\cite{honarvar2020domain}, and domain randomization~\cite{zakharov2019deceptionnet}, arbitrary style transfer~\cite{somavarapu2020frustratingly}, also with Mixup variants~\cite{zhou2021domain}.
2) Empirical and theoretical proofs have shown the effectiveness of learning invariance across domains or aligning the distributions between domains~\cite{zhou2022domain}, such as minimizing maximum mean discrepancy~\cite{li2018domain}, and KL divergence~\cite{li2020domain}. 
3) Researchers have also explored various learning paradigms in the context of DG. \cite{cha2021swad} proposes a stochastic weight averaging densely (SWAD) ensemble algorithm to find flatter minima to avoid overfitting.~\cite{li2022sparse} aligning mixture-of-experts with DG to improve generalization, along with the fusion of self-supervised learning~\cite{kim2021selfreg, zhang2022towards, harary2022unsupervised}. 
Unlike these methods that only focus on domain-invariant patterns, we uncover the invariance across domains by leveraging domain-specific features.

\paragraph{Contrastive Learning.} By maximizing agreement between positive pairs (similar samples) and minimizing agreement between negative pairs (dissimilar samples), contrastive learning (CL) enables the model to capture discriminative features~\cite{chen2020simple, he2020momentum, chen2020improved}. 
Research also shows that more negative samples help improve performance~\cite{wang2020understanding}. 
A few works in DG utilize CL to eliminate domain-specific information from the extracted features of the samples~\cite{li2023cdta, yao2022pcl, jeon2021feature}, as shown in Fig.~\ref{fig:fig2}(a) and (b). 
Our work focuses on utilizing the inherent mutually exclusive features in DG as additional negative samples, making it suitable for leveraging contrastive learning (CL) to exploit information.

\paragraph{Disentanglement Representation Learning.}
This learning strategy aims to identify and disentangle hidden information in data, showing its superiority in model controllability~\cite{liu2022learning}. 
Different from only learning domain-invariant features in DG, disentanglement-based DG methods decompose a feature representation into domain-specific features and domain-invariant features, as shown in Fig.\ref{fig:fig2}(c). Most of them are generative model-based~\cite{qiao2020learning, wang2021variational}, and a few try to decompose component information in network parameters or architecture~\cite{triantafillou2021learning, li2023frequency}. \cite{jo2023poem} also indicates that none of the existing methods are able to identify both the domain-specific and domain-invariant features. 
Different from these methods, our work decouples features by leveraging these two types of features with CL to enhance feature disentanglement.

%-------------------------------------------------------------------------
\section{Method} 
\label{sec:method}

\subsection{Motivation and design}
We start by introducing the motivation of our method before explaining its details. DG aims to learn shared representations on multiple existing source domain datasets to achieve good generalization performance on unseen target domains~\cite{zhou2022domain, wang2022generalizing}. Although deep networks can extract representations from these domains to perform well on training domains, such constraints are still insufficient to guarantee performance. Therefore, we aim to introduce new constraints from the perspective of feature decomposition on latent space representation. Specifically, we require the model to find domain-invariant representations and domain-specific representations during the training process. We believe that if the model can identify strongly correlated features with the domain from a given sample, it can also help find representations related to label information that can cross domains.

Inspired by the content-style disentanglement strategy in disentanglement representation learning and contrastive learning in leveraging sample information in the latent space, we introduce supervised contrastive learning to help decompose domain-specific and domain-invariant representations in the latent space. Specifically, as shown in Fig~\ref{fig:method}, for each sample, we use two feature extractors to extract domain-invariant features and domain-related features, respectively. After that, classifiers are used to calculate prediction errors on class labels and domain labels. Note that each sample only has one class label corresponding to the domain-invariant information. To find the domain-specific information, we manually generate the corresponding domain label to help calculate the error and achieve the goal of training the domain feature extractor.

% \begin{figure}[t]
%     \centering
%     \begin{tikzpicture}
%         % \node at (1.5, 0) (zs) {$\Uparrow\,$Leverage $\mathcal{L}_{dscl}$$\,\Downarrow$};
%         % \node at (3.5, 0) (y) {$\Longrightarrow\, \mathcal{L}_{ce\_dis} \,\Longleftarrow$};
%         \node at (2.5, 1.25) {$\Downarrow$};
%         \node at (2.5, -1.25) {$\Uparrow$};
%         \node at (2.5,0) (dscl) {Leverage $\mathcal{L}_{dscl}$};
%         \node at (5,0) (ce) {$\mathcal{L}_{dscl}$};
%         \node[draw, circle] at (0, 0) (x) {$\mathbf{x}$};
%         \node at (2.5, 2.5) (zs) {$\boldsymbol{z}_s$};
%         \node at (2.5, -2.5) (zv) {$\boldsymbol{z}_v$};
%         \node at (5, 2.5) (yd) {$\hat{y^{\prime}}$};
%         \node at (5, -2.5) (y) {$\hat{y}$};
%         % \node at (-1.75, 3.5) (gdy) {$y$};
%         % \node at (1.75, 3.5) (gdyd) {$y^{\prime}$};
%         \path[->]
%             (x) edge [>=latex] node[above,rotate=45] {$g_s(\cdot)$} (zs)
%             (x) edge [>=latex] node[below,rotate=-45] {$g_v(\cdot)$} (zv)
%             (zs) edge[>=latex] node[above,rotate=0] {$f_s(\cdot)$} (yd)
%             (zv) edge[>=latex] node[below,rotate=0] {$f_v(\cdot)$} (y);
%             % (zs) edge[>=latex] (dscl)
%             % (zv) edge[>=latex] (dscl)
%             % (y) edge[>=latex] (ce)
%             % (yd) edge[>=latex] (ce);
%         % \path[<->]
%         %     (zs) edge [>=latex] node[above,rotate=0] {Leverage $\mathcal{L}_{dscl}$} (zv);
%     \end{tikzpicture}
%     \caption{}
%     \label{fig:framework}
% \end{figure}

\subsection{Preliminary of domain generalization}
First, we introduce the formulation of DG. Let $\mathcal{X}$ be one input data space, and $\mathcal{Y}$ be one output class label space, then one domain is composed of data sampled from the joint distribution $P_{XY}$ on $\mathcal{X}$ and $\mathcal{Y}$, we formulate one domain as $D=\left\{\left({\mathbf{x}}_i, y_i\right)\right\}_{i=1}^N \sim P_{XY}$, where $N$ is the number of data points in this domain, and $\mathbf{x} \in \mathcal{X} \subset \mathbb{R}^d$, $y \in \mathcal{Y} \subset \mathbb{R}$. In DG, there exist multiple source domains $\mathcal{D}=\left\{D^{j}=\left\{\left({\mathbf{x}}_i^{j}, y_i^{j}\right)\right\}_{i=1}^{N_j}\right\}_{j=1}^M$, where $M$ is the number of domains and $N_j$ is the number of data points in $j$-th domain. Note that each domain is individual, thus the distribution of each domain is different: $P_{XY}^{j} \neq P_{XY}^{j^{\prime}}$ when $j \neq j^{\prime}$ and $j, j^{\prime} \in \{1, \ldots, M\}$. Given a test target domain $D_{\mathcal{T}}$ that is unseen during the training phase, the goal of DG is then to learn a generalizable predictive hypothesis $h: \mathcal{X} \rightarrow \mathcal{Y}$ from $\mathcal{D}$ to minimize the prediction error on $D_{\mathcal{T}}$. Note that the target domain also has a distinct distribution thus $P_{XY}^{\mathcal{T}} \neq P_{XY}^{j}, \forall j \in\{1, \ldots, M\}$. The whole optimization for DG can be denoted as follows:
\begin{equation}
\label{eq:dg}
    \min _h \mathbb{E}_{(\mathbf{x}, y) \in \mathcal{D}_{\mathcal{T}}} [\ell\left(h(\mathbf{x}), y\right)],
\end{equation}
where $\mathbb{E}$ is the expectation and $\ell(\cdot, \cdot)$ is the loss function. Specifically, for learning $h$, if using cross-entropy as the loss function, and calculating over multiple source domains, Eq. \ref{eq:dg} can also be written with cost function as:
\begin{equation}
\label{eq:dg_theta}
    \begin{aligned}
         \mathcal{L} &= \mathcal{L}_{ce} + \lambda\ell_{reg},
        \\
        \mathcal{L}_{ce} &= \frac{1}{M} \sum_{j=1}^M \frac{1}{N_j} \sum_{i=1}^{N_j} \ell\left(h(\mathbf{x}_i^{j}), y_i^{j}\right),
    \end{aligned}
\end{equation}
Where $\lambda$ is a trade-off factor and $R(\cdot)$ is a regularization term to prevent overfitting, which could be omitted for simplicity.
% \begin{equation}
% \label{eq:dg_theta}
%     \begin{aligned}
%         & \hat{\theta}=\underset{\theta}{\arg \min } \frac{1}{M} \sum_{j=1}^M \mathcal{L}_j(\theta)+\lambda R(\theta), 
%         \\
%         \textit{where } & \mathcal{L}_j(\theta)=\frac{1}{N_j} \sum_{i=1}^{N_j} \ell\left(h\left(\mathbf{x}_i^{j} ; \theta\right), y_i^{j}\right),
%     \end{aligned}
% \end{equation}

\subsection{Disentanglement domain generalization}
We decompose the prediction hypothesis $h$ into representation generator $g$ and classifier $f$ as $h=f \circ g$, and as previously described, we generate domain labels for disentangled domain representation, which we denote as $y^{\prime} \in \mathcal{Y^{\prime}} \subset \mathbb{R}$, and $\mathcal{Y} \cap \mathcal{Y^{\prime}} = \emptyset$. Since disentanglement-based DG methods decompose a feature representation into domain-invariant and domain-specific features, we also calculate the prediction error on domain features with generated domain labels. Thus the optimization goal turns out to be:
\begin{equation}
\resizebox{\columnwidth}{!}{%
    $ \min _h \mathbb{E}_{(\mathbf{x}, y) \in \mathcal{D}_{\mathcal{T}}} [\ell(f_v(g_v(\mathbf{x})), y)] +
    \mathbb{E}_{(\mathbf{x}, y) \in \mathcal{D}_{\mathcal{T}}} [\ell(f_s(g_s(\mathbf{x})), y^{\prime})] $,
}%
\end{equation}
where $g_v$ and $g_s$ indicate the domain-invariant and domain-specific feature representation generator, respectively. Besides, Eq.~\ref{eq:dg_theta} turns to:
\begin{equation}
    \begin{aligned}
        \mathcal{L} & =\mathcal{L}_{ce\_dis}+\lambda \ell_{r e g} \\
        \textit { where } \mathcal{L}_{ce\_dis} & =\frac{1}{M} \sum_{j=1}^M \frac{1}{N_j} \sum_{i=1}^{N_j}\biggl(\ell\left(f_v(g_v(\mathbf{x}_i^j)), y_i^j\right)\biggl. \\
        &\biggl. \quad + \ell\left(f_s(g_s(\mathbf{x}_i^j)), (y^{\prime})_i^j)\right)\biggl)
    \end{aligned}
\end{equation}
After feature disentanglement, each sample naturally exhibits two mutually exclusive representations. To further enhance this mutual exclusivity and facilitate decoupling, we employ contrastive learning, which has shown promising results in leveraging samples in the latent space by repelling negative samples and attracting positive samples.

\subsection{DG-specific feature disentanglement with contrastive learning}
With DG-specific disentanglement, we now have domain-invariant feature $g_v(\mathbf{x})$ and domain-specific feature $g_s(\mathbf{x})$ for each sample, to better improve the constrain for disentanglement, we map these two extracted features into latent spaces where we employ contrastive learning. By contrasting positive pairs (similar samples) against negative pairs (dissimilar samples), the model is incentivized to disentangle the underlying factors that distinguish different samples (universal patterns across domains or categories). To fully leverage all the obtained features, we have devised a unifying framework that takes into account the pairwise relationships among features of arbitrary types, as shown in Fig.~\ref{fig:method}. Since we have the ground-truth class label and domain label, and these labels do not share label space, the mapping leads to two ways:

\begin{itemize}
    \item \textbf{Mixed label space mapping}, which we named as \textit{CDDG\_comb}. This variant maps all domain features and class features of each sample into the same latent space, where there are a total of $|\left\{\mathcal{Y} \cup \mathcal{Y}^{\prime}\right\}|$ classes, i.e., we mix the extracted class feature samples and domain feature samples, and simultaneously \textit{\textbf{combine}} their respective label spaces. For example, PACS has 4 domains and 7 categories. Thus the mixed label space has a total of 11 classes. This mapping introduces a strong constraint: finding a latent space that satisfies the mixing of label space and leveraging samples on this space based on contrastive learning.
    \item\ \textbf{Independent label space mapping}, which we named as \textit{CDDG\_ind}. We look for two individual feature spaces for class features and domain features, where one type of feature is mapped into one feature space while the left is mapped as an additional class, leading to a new label space of $\left\{\mathcal{Y}+1\right\}$ for class label space and $\left\{\mathcal{Y}^{\prime}+1\right\}$ for domain label space, i.e., we mix the samples from two encoders, but still left the respective label spaces \textbf{\textit{independent}}. Note that though this operation is relatively simple (take all other features as a whole and align an additional class to them, e.g., mapping one type of feature into the other feature space, and aligning additional fake labels for them), this may introduce noise during positive and negative pair selection, since the extra samples have multiple classes, but they are aligned as one give class.
\end{itemize}

We start by giving loss definitions of supervised contrastive learning. \cite{khosla2020supervised} re-define InfoNCE~\cite{chen2020simple} loss to a supervised version to incorporate label information, for a batch with augmented samples of $I \equiv\{1 \ldots 2 N\}$, the loss of supervised contrastive learning is:
\begin{equation}
\label{eq:scl}
    \mathcal{L}_{scl}=\sum_{i \in I} \frac{-1}{|P(i)|} \sum_{p \in P(i)} \log \frac{\exp \left(\boldsymbol{z}_i \cdot \boldsymbol{z}_p / \tau\right)}{\sum_{a \in A(i)} \exp \left(\boldsymbol{z}_i \cdot \boldsymbol{z}_a / \tau\right)}
\end{equation}
where $\boldsymbol{z}$ stands for extracted features, the $\cdot$ denotes the inner (dot) product, $\tau \in \mathcal{R}^{+}$ is a scalar temperature parameter. $P(i) \equiv\left\{p \in A(i): \boldsymbol{y}_p=\boldsymbol{y}_i\right\}$, $|P(i)|$ is its cardinality, and $A(i) \equiv I \backslash\{i\}$. In eq.~\ref{eq:scl}, the index $\boldsymbol{i}$ is known as the \textit{anchor}, and the other index $\boldsymbol{p}$ in the numerator stands for the \textit{positive} sample index, and the other indices ($\left\{A(i) \backslash P(i)\right\}$) are \textit{negative}.

Combining previous symbols, we have a batch of $I \equiv\{1 \ldots 2 N\}$, and after feature extraction, we have $S \equiv\{1 \ldots 2 N\}$ of domain feature samples and $V \equiv\{1 \ldots 2 N\}$ of class feature samples, and in a total of $I \equiv\{1 \ldots 4 N\}$ samples. The loss of DG-specific feature disentanglement with contrastive learning (CDDG) can be written as:
\begin{equation}
    \mathcal{L}_{dscl} = \begin{cases}\mathcal{L}_{dscl\_comb}, & \text { if } y \in \left\{\mathcal{Y} \cup \mathcal{{Y}^{\prime}}\right\} \\ \mathcal{L}_{dscl\_ind}, & \text { if } y \in \left\{\mathcal{Y}+1\right\} \textit{or} \left\{\mathcal{Y}^{\prime}+1\right\}\end{cases}
\end{equation}
where $\mathcal{L}_{dscl\_comb}$ leads to a mixed calculation when mapping category label space and domain label space into one; and $\mathcal{L}_{dscl\_ind}$ leads to an independent calculation when two label spaces are separate:
\begin{equation}
    \begin{aligned}
        & \mathcal{L}_{dscl\_comb}= \\
        & \sum_{i \in I} \frac{-1}{|P(i)|} \sum_{p \in P(i)} \log \frac{\exp \left(g(i) \cdot g(p) / \tau\right)}{\sum_{a \in A(i)} \exp \left(g(i) \cdot g(a) / \tau\right)}
    \end{aligned}
\end{equation}
where $g(i)$ and $g(p)$ stand for arbitrary features, while for $\mathcal{L}_{dscl\_ind}$, $g(i)$ is separated to $g_s(s)$ for features from domain-specific extractor and $g_v(v)$ for domain-invariant extractor as:
\begin{equation}
\resizebox{\columnwidth}{!}{%
    $\begin{aligned}
        & \mathcal{L}_{dscl\_ind}= \\
        & \sum_{s \in S} \frac{-1}{|P(s)|} \sum_{p \in P(s)} \log \frac{\exp \left(g_s(s) \cdot g_s(p) / \tau\right)}{\sum_{a \in \left\{S(s) \cup V\right\}} \exp \left(g_s(s) \cdot g_s(a) / \tau\right)}+\\
        & \sum_{v \in V} \frac{-1}{|P(v)|} \sum_{p \in P(v)} \log \frac{\exp \left(g_v(v) \cdot g_v(p) / \tau\right)}{\sum_{a \in \left\{V(v) \cup S\right\}} \exp \left(g_v(v) \cdot g_v(a) / \tau\right)}
    \end{aligned}$
}%
\end{equation}
where $S(s) \equiv S \backslash\{s\}$ and $V(v) \equiv V \backslash\{v\}$, and $P(s) \equiv \left\{p \in S(s): \boldsymbol{y}_p=\boldsymbol{y}_s\right\}$ is the set of indices of all \textit{positives} (with same domain label) in domain feature sample set, $P(v) \equiv \left\{p \in V(v) : \boldsymbol{y}_p=\boldsymbol{y}_v\right\}$ is the set of indices of all \textit{positives} (with same class label) in the class feature sample set. Algorithm~\ref{alg:cddg} summarizes the proposed method.

\begin{algorithm}[!ht]
\DontPrintSemicolon
  
  \KwInput{batch size $N$, weight factor $\alpha$, structure of $g_v, g_s$, $f_v, f_s$.}
  \KwOutput{domain feature encoder $g_s(\cdot)$, class feature encoder $g_v(\cdot)$, domain classifier $f_s$, class classifier $f_v$.}
  \KwData{Training data $\mathbf{x} \in \mathcal{X}, |\mathcal{X}|=2N$, sampled from training domains with augmented views.}
  \For{all $k \in {1, \ldots, N}$}
    {
        $g_s(\mathbf{x}_k)$, $g_v(\mathbf{x}_k)$ \tcp{extracted features after feeding to encoders}
        $\hat{y}^{\prime}_k = f_s(g_s(\mathbf{x}_k))$,  
        $\hat{y}_k = f_v(g_v(\mathbf{x}_k)) $ \tcp{predicted labels}
    }
  \tcc{CDDG\_comb}
  \If{CDDG\_comb}
    {
        $\mathcal{L}$ = $\mathcal{L}_{ce\_dis} + \alpha\mathcal{L}_{dscl\_comb}$
    }
  \tcc{CDDG\_ind}
  \ElseIf{CDDG\_ind}
    {
        $\mathcal{L}$ = $\mathcal{L}_{ce\_dis} + \alpha\mathcal{L}_{dscl\_ind}$
    }
  update networks $g_v, g_s$, $f_v, f_s$ to minize $\mathcal{L}$
\caption{Algorithm of \textit{CDDG}}\label{alg:cddg}
\end{algorithm}

%-------------------------------------------------------------------------
\section{Experiments}
In this section, we evaluate our methods by measuring image classification accuracy on four common DG image classification datasets. Our work is built mainly on DomainBed~\cite{gulrajani2021search}, which is a DG benchmark with famous state-of-the-art works in recent years. 

\subsection{Dataset details}
% \begin{table}[!ht]
%     \begin{center}
%     \resizebox{\columnwidth}{!}{%
%         \begin{tabular}{ccccc}
%             \toprule[1.2pt]
%             \textbf{Dataset} & \textbf{\# Classes} & \textbf{\# Domains} & \textbf{Details} & \textbf{\# Images} \\ \midrule
%                 PACS & 7 & 4 & {P\tiny hoto}, {A\tiny rt}, {C\tiny artoon}, {S\tiny ketch} & 9,991 \\ \midrule
%                 VLCS & 5 & 4 & {V\tiny OC2007}, {L\tiny abelMe}, {C\tiny altech101}, {S\tiny UN09} & 10,729 \\ \midrule
%                 Office-Home & 65 & 4 & {A\tiny rt}, {C\tiny lipart}, {P\tiny roduct}, {R\tiny eal} & 15,588 \\ \midrule
%                 DomainNet & 345 & 6 & {C\tiny lipart}, {I\tiny nfograph}, {P\tiny ainting}, {Q\tiny uickdraw}, {R\tiny eal}, {S\tiny ketch} & 586,575 \\ 
%             \bottomrule[1.2pt]
%         \end{tabular}
%     }
%     \end{center}
%     \caption{Details of datasets for evaluation.}
%     \label{tab:dataset}
% \end{table}

\begin{table*}[!ht]
\begin{center}
\resizebox{\linewidth}{!}{%
\begin{tabular}{cccccccccccc}
\toprule[1.2pt]

\multirow{2}{*}{\textbf{Group}} & \multirow{2}{*}{\textbf{Algorithms}} & \multicolumn{2}{c}{\textbf{PACS}} & \multicolumn{2}{c}{\textbf{VLCS}} & \multicolumn{2}{c}{\textbf{Office-Home}} & \multicolumn{2}{c}{\textbf{DomainNet}}  & \multicolumn{2}{c}{\textbf{Avg.}} \\ \cmidrule(lr){3-4} \cmidrule(lr){5-6} \cmidrule(lr){7-8} \cmidrule(lr){9-10} \cmidrule(lr){11-12}
  & & \textit{TDVS} & \textit{Oracle} & \textit{TDVS} & \textit{Oracle} & \textit{TDVS} & \textit{Oracle} & \textit{TDVS} & \textit{Oracle} & \textit{TDVS} & \textit{Oracle} \\ \midrule

Baseline                      & ERM        & 
85.5 $\pm$ 0.2 & 86.7 $\pm$ 0.3 & 77.5 $\pm$ 0.4 & 77.6 $\pm$ 0.3 & 
66.5 $\pm$ 0.3 & 66.4 $\pm$ 0.5 & 40.9 $\pm$ 0.1 & 41.3 $\pm$ 0.1 & 67.6 & 68.0 \\ 
\midrule
\multirow{2}{*}{Optimization} & GroupDRO   & 
84.4 $\pm$ 0.8 & 87.1 $\pm$ 0.1 & 76.7 $\pm$ 0.6 & 77.4 $\pm$ 0.5 &
66.0 $\pm$ 0.7 & 66.2 $\pm$ 0.6 & 33.3 $\pm$ 0.2 & 33.4 $\pm$ 0.3 & 65.1 & 66.0 \\
                              & MLDG       & 
84.9 $\pm$ 1.0 & 86.8 $\pm$ 0.4 & 77.2 $\pm$ 0.4 & 77.5 $\pm$ 0.1 &
66.8 $\pm$ 0.6 & 66.6 $\pm$ 0.3 & 41.2 $\pm$ 0.1 & 41.6 $\pm$ 0.1 & 67.5 & 68.1       \\ \midrule
\multirow{4}{*}{Augmentation} & VREx       &
84.9 $\pm$ 0.6 & 87.2 $\pm$ 0.6 & 78.3 $\pm$ 0.2 & 78.1 $\pm$ 0.2 &
66.4 $\pm$ 0.6 & 65.7 $\pm$ 0.3 & 33.6 $\pm$ 2.9 & 30.1 $\pm$ 3.7 & 65.8 & 65.3 \\
                              & ARM        &
85.1 $\pm$ 0.4 & 85.8 $\pm$ 0.2 & 77.6 $\pm$ 0.3 & 77.8 $\pm$ 0.3 &
64.8 $\pm$ 0.3 & 64.8 $\pm$ 0.4 & 35.5 $\pm$ 0.2 & 36.0 $\pm$ 0.2 & 65.8 & 66.1 \\
                              & MixUp      & 
84.6 $\pm$ 0.6 & 86.8 $\pm$ 0.3 & 77.4 $\pm$ 0.6 & 78.1 $\pm$ 0.3 &
68.1 $\pm$ 0.3 & 68.0 $\pm$ 0.2 & 39.2 $\pm$ 0.1 & 39.6 $\pm$ 0.1 & 67.3 & 68.1 \\
                              & SagNet     &
86.3 $\pm$ 0.2 & 86.4 $\pm$ 0.4 & 77.8 $\pm$ 0.5 & 77.6 $\pm$ 0.1 &
68.1 $\pm$ 0.1 & 67.5 $\pm$ 0.2 & 40.3 $\pm$ 0.1 & 40.8 $\pm$ 0.2 &
68.1 & 68.1 \\ 
\midrule
\multirow{6}{*}{Invariant}    & MMD        &
84.6 $\pm$ 0.5 & 87.2 $\pm$ 0.1 & 77.5 $\pm$ 0.9 & 77.9 $\pm$ 0.1 &
66.3 $\pm$ 0.1 & 66.2 $\pm$ 0.3 & 23.4 $\pm$ 9.5 & 23.5 $\pm$ 9.4 & 63.0 & 63.7 \\
                              & IRM        & 
83.5 $\pm$ 0.8 & 84.5 $\pm$ 1.1 & 78.5 $\pm$ 0.5 & 76.9 $\pm$ 0.6 &
64.3 $\pm$ 2.2 & 63.0 $\pm$ 2.7 & 33.9 $\pm$ 2.8 & 28.0 $\pm$ 5.1 & 65.1 & 63.1 \\
                              & CDANN      &
82.6 $\pm$ 0.9 & 85.8 $\pm$ 0.8 & 77.5 $\pm$ 0.1 & 79.9 $\pm$ 0.2 &
65.8 $\pm$ 1.3 & 65.3 $\pm$ 0.5 & 38.3 $\pm$ 0.3 & 38.5 $\pm$ 0.2 & 66.1 & 67.4 \\
                              & DANN       &
83.6 $\pm$ 0.4 & 85.2 $\pm$ 0.2 & 78.6 $\pm$ 0.4 & 79.7 $\pm$ 0.5 &
65.9 $\pm$ 0.6 & 65.3 $\pm$ 0.8 & 38.3 $\pm$ 0.1 & 38.3 $\pm$ 0.1 & 66.6 & 67.1 \\
                              & RSC        &
85.2 $\pm$ 0.9 & 86.2 $\pm$ 0.5 & 77.1 $\pm$ 0.5 & 77.8 $\pm$ 0.6 &
65.5 $\pm$ 0.9 & 66.5 $\pm$ 0.6 & 38.9 $\pm$ 0.5 & 38.9 $\pm$ 0.6 & 66.7 & 67.4 \\ 
                              & MTL        &
84.6 $\pm$ 0.5 & 86.7 $\pm$ 0.2 & 77.2 $\pm$ 0.4 & 77.7 $\pm$ 0.5 &
66.4 $\pm$ 0.5 & 66.5 $\pm$ 0.4 & 40.6 $\pm$ 0.1 & 40.8 $\pm$ 0.1 & 67.2 & 67.9 \\
                              & CORAL      &
86.2 $\pm$ 0.3 & 87.1 $\pm$ 0.5 & 78.8 $\pm$ 0.6 & 77.7 $\pm$ 0.2 &
\textbf{68.7 $\pm$ 0.3} & \textbf{68.4 $\pm$ 0.2} & 41.5 $\pm$ 0.1 & 41.8 $\pm$ 0.1 & 68.8 & 68.8 \\
\midrule
\multirow{2}{*}{Disentanglement} & POEM   & 
86.7 $\pm$ 0.2 & - & 79.2 $\pm$ 0.6 & - & 68.0 $\pm$ 0.2 & - & 44.0 $\pm$ 0.0 & - & 69.5 & - \\
% Ours               & \textit{\textbf{CDDG}}  & 
% \textbf{87.5 $\pm$ 0.5} & \textbf{88.7 $\pm$ 0.4} & \textbf{80.2 $\pm$ 0.2} & \textbf{81.0 $\pm$ 0.2} & 68.1 $\pm$ 0.7 & 67.2 $\pm$ 0.4 & \textbf{44.6 $\pm$ 0.2} & \textbf{43.9 $\pm$ 0.2} & \textbf{70.1} & \textbf{70.2} \\
  & \textit{\textbf{CDDG} (Ours)}  & 
\textbf{87.5 $\pm$ 0.5} & \textbf{88.7 $\pm$ 0.4} & \textbf{80.2 $\pm$ 0.2} & \textbf{81.0 $\pm$ 0.2} & 68.1 $\pm$ 0.7 & 67.2 $\pm$ 0.4 & \textbf{44.6 $\pm$ 0.2} & \textbf{43.9 $\pm$ 0.2} & \textbf{70.1} & \textbf{70.2} \\
\bottomrule[1.2pt]
\end{tabular}%
}
\end{center} 
\caption{Test accuracy (\%) with state-of-the-art methods (divided into five categories according to algorithm details) on four datasets from DomainBed benchmark. \textit{TDVS} stands for one model selection method of the training-domain validation set, while \textit{Oracle} represents the test-domain validation set. The best numbers are in \textbf{bold}.}
\label{tab:allresult}
\end{table*}

\begin{table}[!ht]
    \begin{center}
        \resizebox{\columnwidth}{!}{%
            \begin{tabular}{c|c|ccccc}
            \toprule[1.2pt]
            \textbf{Model} & \multirow{2}{*}{\textbf{Ablation}} & \multirow{2}{*}{\textbf{PACS}} & \multirow{2}{*}{\textbf{VLCS}} & \multirow{2}{*}{\textbf{Office-Home}} & \multirow{2}{*}{\textbf{DomainNet}} & \multirow{2}{*}{\textbf{Avg.}} \\ 
            \textbf{selection} & & & & & & \\
            \midrule
            \multirow{4}{*}{\textit{TDVS}} & \textit{CDDG}   & \textbf{87.5\small$\pm$0.5} & \textbf{80.2\small$\pm$0.2} & \textbf{68.1\small$\pm$0.7} & \textbf{44.6\small$\pm$0.2} & \textbf{70.1} \\ 
            \cmidrule{2-7}
            & w/ $\mathcal{L}_{dscl\_ind}$ & 86.3\small$\pm$0.3 & 75.2\small$\pm$1.5 & 64.8\small$\pm$0.5 & 41.8\small$\pm$0.8 & 67.0 \\
            & w/o $\mathcal{L}_{dscl\_comb}$  & 84.5\small$\pm$0.7 & 78.3\small$\pm$0.5 & 66.3\small$\pm$0.3 & 43.5\small$\pm$0.7 & 68.2 \\
            & w/o $\mathcal{L}_{ce\_dis}$  & 85.8\small$\pm$0.2 & 77.7\small$\pm$0.8 & 67.3\small$\pm$0.5 & 42.9\small$\pm$0.3 & 68.4 \\ \midrule
            \multirow{4}{*}{\textit{Oracle}} & \textit{CDDG}   & \textbf{88.7\small$\pm$0.4} & \textbf{81.0\small$\pm$0.2} & 67.2\small$\pm$0.4 & \textbf{43.9\small$\pm$0.2} & \textbf{70.2}  \\
            \cmidrule{2-7}
            & w/ $\mathcal{L}_{dscl\_ind}$ & 87.9\small$\pm$0.3 & 77.8\small$\pm$2.1 & \textbf{67.3\small$\pm$0.1} & 43.0\small$\pm$0.5 & 69.0 \\
            & w/o $\mathcal{L}_{dscl\_comb}$  & 85.7\small$\pm$1.5 & 78.9\small$\pm$0.3 & 65.5\small$\pm$0.2 & 42.5\small$\pm$0.3 & 67.3 \\
            & w/o $\mathcal{L}_{ce\_dis}$  & 86.9\small$\pm$0.2 & 79.2\small$\pm$0.3 & 66.1\small$\pm$0.3 & 43.3\small$\pm$0.1 & 68.9\\ 
            \bottomrule[1.2pt]
            \end{tabular}%
        }
    \end{center}
    \caption{Ablation study of the proposed method. The best result is highlighted in \textbf{bold}. }
    \label{tab:ablation}
\end{table}

\textbf{PACS} is a dataset with 9991 images in total and containing four domains. Examples can be seen in Fig.~\ref{fig:intro}. Each domain contains seven categories. \textbf{VLCS} is another common DG dataset comprising photographic domains of VOC2007, LabelMe, Caltech101, and SUN09. \textbf{Office-Home} has four domains with 65 categories and contains 15,588 images. We also evaluate on \textbf{DomainNet}, a large-scale dataset with six domains, 345 categories, and 586,575 images. Samples in PACS, Office-Home, and DomainNet are with style shifts across domains, while in VLCS, the main shift is mainly caused by object viewpoint or environment changes.

\subsection{DomainBed settings and model selection}
Following~\cite{gulrajani2021search}, all \textbf{feature extractor} used in this work, including $g_v(\cdot)$ and $g_s(\cdot)$, are ResNet-50~\cite{he2016deep}. \textbf{Data augmentation} plays an important role in DG since it can somehow approximate variations in domains. Following~\cite{gulrajani2021search}, we employ simple standard image data augmentation in this work, and no additional augmentation methods are involved. As for \textbf{data split}, we follow the original protocol in DomainBed, which splits each source domain into one training set with 80\% data and one validation set with left 20\%. \textbf{Metrics}. The criterion used in this work is leave-one-domain-out, which iteratively chooses one domain as the unseen target domain for evaluation while the left domains are taken as the training domains. As described in~\cite{gulrajani2021search}, the absence of details of \textbf{model selection} brings confusion for comparison with other methods. We list two commonly used model selection methods here: Training-domain validation set (TDVS), which samples validation data from \textit{seen} training domains; Test-domain validation set (Oracle), where validation data are from \textit{unseen} target domain. Following~\cite{gulrajani2021search}, we conduct a random search for each algorithm and test environment, and report our entire experimental results three times, ensuring every random choice makes all settings, including hyperparameters, and data split anew.

\subsection{Main results}
We describe in this section the complete evaluation results with DomainBed~\cite{gulrajani2021search}. The comparison methods we list here are categorized into five groups: the baseline for ERM~\cite{vapnik1999overview}; Optimization-based methods of GroupDRO~\cite{sagawa2019distributionally} and MLDG~\cite{li2018learning}; Augmentation-based methods of MixUp~\cite{yan2020improve}, ARM~\cite{zhang2021adaptive}, VREx~\cite{krueger2021out}, and SagNet~\cite{nam2021reducing}; The mainstream invariant representation learning methods including IRM~\cite{arjovsky2019invariant}, MMD~\cite{li2018domain}, DANN~\cite{ganin2016domain}, CDANN~\cite{li2018deep}, CORAL~\cite{sun2016deep}, and RSC~\cite{huang2020self}; and feature disentanglement of POEM~\cite{jo2023poem}. Note that ensemble learning provides a significant performance improvement, to make a fair comparison, methods including SWAD~\cite{cha2021swad}, SWAD-based methods, e.g, PCL~\cite{yao2022pcl}, MIRO~\cite{cha2022domain}, and POEM variants~\cite{jo2023poem} are not listed here. It should be noted that our methods can also serve as a recipe for ensemble learning for better results. Results in Table.~\ref{tab:allresult} indicate that our proposed method \textit{CDDG} outperforms all other methods in average. While CORAL achieves the best result in the dataset Office-Home, our method reports state-of-the-art results in the other three datasets. Note that algorithms such as POEM are not listed as we only report methods that have both TDVS and Oracle results here, and the results of \textit{CDDG} we report are from \textit{CDDG\_comb} since this variant demonstrates a better performance than \textit{CDDG\_ind}, and thus we take \textit{CDDG\_ind} variant as a part of ablation study of our method.

\subsection{Ablation study}\label{sec: ablation study}

\paragraph{Ablation on CDDG variants.} CDDG has two variants of \textit{CDDG\_comb} and \textit{CDDG\_ind}, leading to two different mapping ways. The first row in Table.~\ref{tab:ablation} represents for the results of \textit{CDDG\_comb} while the second is for \textit{CDDG\_ind}. In most cases, it is obvious that \textit{CDDG\_comb} outperforms \textit{CDDG\_ind}. The reason we suppose is that the simple expansion of negative samples in \textit{CDDG\_ind} brings additional noise into the training, since this operation ignores the correct label information in the additional samples, either the ground-truth class label information or the added domain label information.

\begin{figure}[!t]
\begin{center}
   \includegraphics[width=1.0\linewidth]{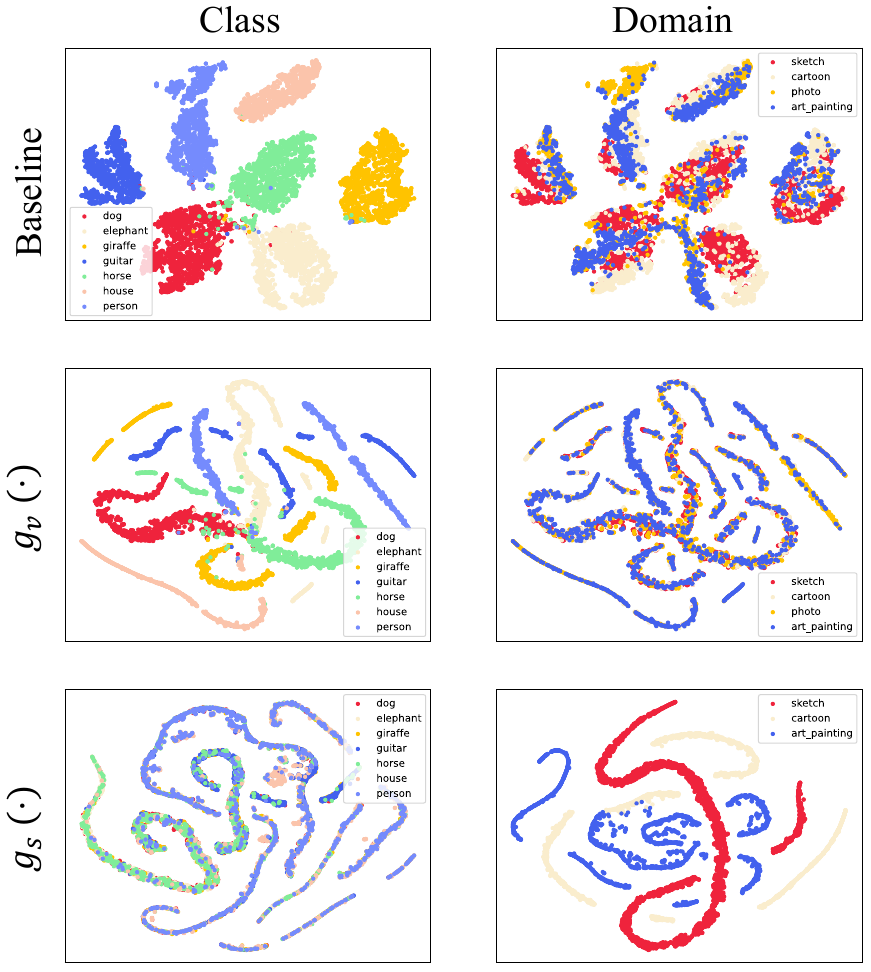}
\end{center}
   \caption{Visualization by t-SNE of PACS dataset for the baseline of ERM, class feature extractor $g_v(\cdot)$ and domain feature extractor $g_s(\cdot)$ of our method \textit{CDDG}. The column name represents the clustering target. $g_v(\cdot)$ demonstrates strong category classification performance, while $g_s(\cdot)$ is capable of classifying domains. Best viewed in color.}
\label{fig:tsne}
\end{figure}

\paragraph{Effect of disentanglement.} First, we evaluate the impact of disentanglement by creating a variant of decoupling based on the ERM baseline, i.e. an additional encoder is set up to extract domain features and the original encoder is used to extract category features only. 
Domain labels are used to constrain the extraction targets of the domain feature extractor. Note that compared to the full \textit{CDDG}, we have not added $\mathcal{L}\_dscl$, so the only constraints on this method are the label information and the domain label information on the features, which is a relatively weak decoupling approach. 
For a fair comparison, we conduct three trials of experiments following the same setup as \textit{CDDG}, and the results can be seen in Table. \ref{tab:ablation}. 
The row \textit{w/o $\mathcal{L}_{ce\_dis}$} shows that if using only disentanglement, there is a significant drop in performance on all datasets in both model selection methods.

\paragraph{Effect of contrastive learning.} We then evaluate the effect of using only contrast learning without introducing disentanglement, again based on ERM baseline, with the difference that only one encoder is used to extract features, but we augment the data samples to calculate the supervised contrastive learning loss. 
The results illustrate that using $\mathcal{L}_{dscl}$ alone shows a significant drop.

\begin{figure}[!t]
\begin{center}
   \includegraphics[width=1.0\linewidth]{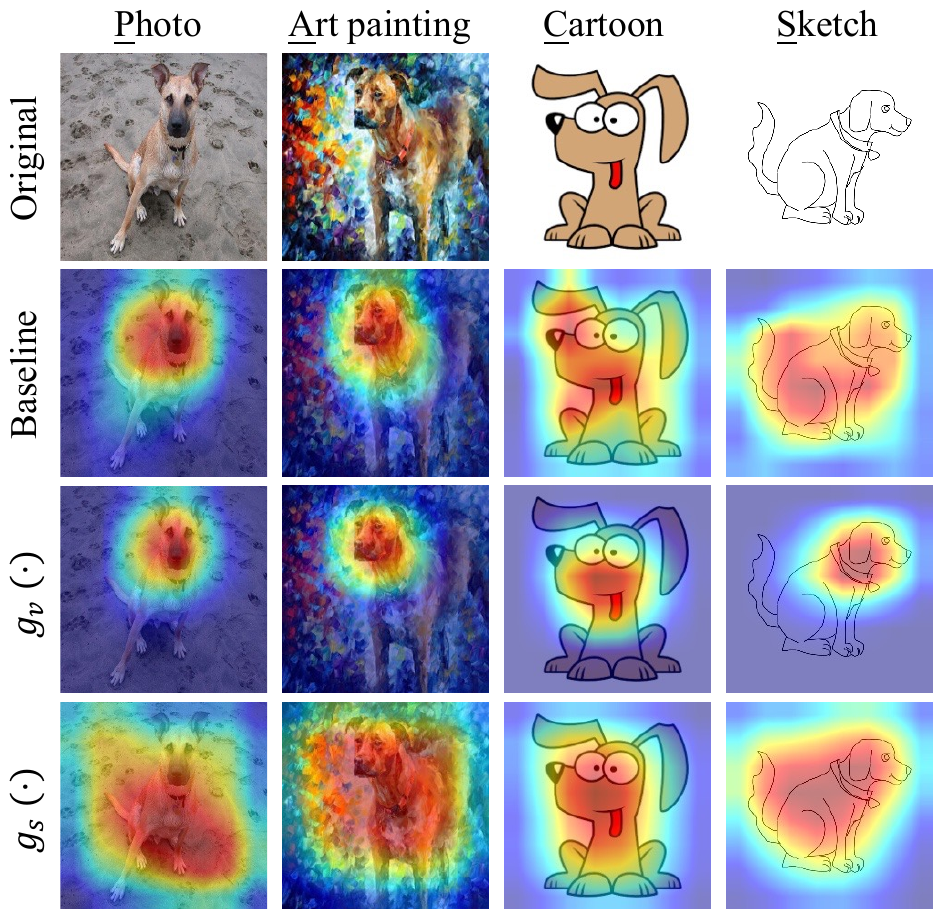}
\end{center}
   \caption{Areas of interest of models from PACS dataset. The first row is the original images with dog class, and the second is the baseline activation maps. The third and fourth rows are for two encoders from our method. Both of them, as designed, respectively prioritize the attention to category information and domain information.}
\label{fig:gradcam}
\end{figure}

\subsection{Additional Evaluation}
To further investigate the feasibility of our proposed methods, we use t-SNE~\cite{van2008visualizing} to plot the clustered sample distribution in latent space. We also employ GradCAM~\cite{selvaraju2017grad} to visualize the areas of interest of models in images.

As shown in Fig.~\ref{fig:tsne}, the baseline model (ERM-based ResNet-50) has a strong initialization ability to categorize samples into groups, but with unsatisfactory domain recognition performance. While there are few samples in the middle that are not well identified, our method (the first column of the row $g_v(\cdot)$ as the class feature encoder of our method \textit{CDDG}) shows a better classification performance. Since we introduce domain information in our method, with domain feature extractor $g_s(\cdot)$, the domains are supposed to be classified. As seen in the second column of the row $g_s(\cdot)$, it is obvious that our method has a strong classification ability of domains since we introduce domain label information as the constraints and extra negative samples of class features during the training phase. We also evaluate the ability to classify domains with $g_v(\cdot)$ and the ability to classify classes with $g_s(\cdot)$, which in our design should be worse, as we added constraints in the opposite direction when training these two encoders. From the domain column row $g_v(\cdot)$ and class column row $g_s(\cdot)$, the results are as we expected, that these two feature extractors have no ability to classify classes (for $g_s(\cdot)$) and domains (for $g_v(\cdot)$).

As shown in Fig.~\ref{fig:gradcam}, the activation maps of vanilla ResNet-50 for the images are mainly focused on those regions related to the class labels, while the class feature extractor $g_v(\cdot)$ in our method has a more class-focused effect, i.e., the activation region is more focused on the key facial features of the dog category. As seen in the fourth row, the domain feature extractor has a broader region of activation maps to encompass more domain details. The differences between vanilla and $g_s(\cdot)$ in cartoon and sketch show that the domain feature extractor is more concerned with edge lines specific to the cartoon and sketch domains, rather than information related to class labels. As for photo and art painting images, $g_s(\cdot)$ tends to seek natural scene and painting features.

%-------------------------------------------------------------------------
\section{Conclusion}
In this paper, we propose CDDG to tackle the domain generalization problem from a novel feature disentanglement perspective with contrastive learning. The direct decoupling of objectives is insufficient to bring about sufficient feature representation capabilities and is prone to falling into local optima. However, the effects brought by contrastive learning can effectively compensate for this limitation, making the overall learning process more stable and discriminative, leading to a promising fusion of a DG-specific contrastive-based disentanglement framework. Empirically, we achieve state-of-the-art performance on various benchmarks and also analyze the benefits of introducing contrastive uniformity with visualization evaluations. We expect our work to provide inspiration for learning DG-specific feature structures in the context of feature decoupling.

{\small
\bibliographystyle{ieee_fullname}
\bibliography{egbib}
}

\end{document}